\documentclass[11pt,letterpaper]{article}
\usepackage{emnlp2017}
\usepackage{booktabs}
\usepackage{enumitem}
\usepackage{times}
\usepackage{adjustbox}
\usepackage{array}
\usepackage[utf8]{inputenc}
\usepackage{amsmath, amssymb, latexsym}
 
\usepackage{tikz}
\usetikzlibrary{decorations.pathreplacing}
\usetikzlibrary{fadings}

\emnlpfinalcopy

\def\emnlppaperid{***}

\title{Sentence-level quality estimation by predicting HTER as a multi-component
metric}

\author{Eleftherios Avramidis \\
	German Research Center for Artificial Intelligence (DFKI) \\
	Language Technology Lab, Berlin, Germany \\
  {\tt eleftherios.avramidis@dfki.de}}

\date{2017-07-14}

\newcolumntype{R}[2]{%
    >{\adjustbox{angle=#1,lap=\width-(#2)}\bgroup}%
    l%
    <{\egroup}%
}
\begin{document}

\maketitle

\begin{abstract}
This submission investigates alternative machine learning models for predicting
the HTER score on the sentence level. Instead of directly predicting the HTER
score, we suggest a model that jointly predicts the amount of the 4 distinct
post-editing operations, which are then used to calculate the HTER score. This
also gives the possibility to correct invalid (e.g.~negative) predicted values
prior to the calculation of the HTER score. Without any feature exploration, a
multi-layer perceptron with 4 outputs yields small but significant improvements
over the baseline.
%, but no positive results are shown on the test set.
\end{abstract}

\section{Introduction}

Quality Estimation (QE) is the evaluation method that aims at employing machine
learning in order to predict some measure of quality given a Machine Translation
(MT) output \cite{Blatz:2004:CEM:1220355.1220401}. A commonly-used subtask of QE
refers to the learning of automatic metrics. These metrics produce a continuous
score based on the comparison between the MT output and a reference translation.
When the reference is a minimal post-edition of the MT output, the quality score
produced is intuitively more objective and robust as compared to other QE
subtasks, where the quality score is assigned directly by the annotators. In
that case, the score is a direct reflection of the changes that need to take
place in order to fix the translation. HTER
\cite{Snover:2009:FAH:1626431.1626480} is the most commonly used metric as it
directly represents the least required post-editing effort.

In order to predict the results of an automatic metric, QE approaches use
machine learning to predict a model that associates a feature vector with the
single quality score. In this case the statistical model treats the automatic
metric as a black box, in the sense that no particular knowledge about the exact
calculation of the quality score is explicitly included in the model.

In this submission we aim to partially break this black-box. We explore the idea
of creating a QE model that does not directly predict the single HTER score, but
it jointly predicts the 4 components of the metric, which are later used for
computing the single score. This way, the structure of the model can be aware of
the distinct factors that comprise the final quality score and also potentially
learn the interactions between them. Hence, the focus of this submission will
remain on machine learning and there will not be exploration in terms of
features. In Section~\ref{sec:previous} we briefly introduce previous work, in
Section~\ref{sec:methods} we provide details about the method, whereas the
experiment results are given in Section \ref{sec:experiment}. In Section
\ref{sec:submission} we describe the models submitted at the shared-task and we
explain why they differ from our best models. Finally, in Section
\ref{sec:conclusions} we present the conclusions and some ideas for future work.

\section{Previous work}
\label{sec:previous}

The prediction of HTER first appeared as a means to estimate post-editing effort
\cite{Specia2010}. Bypassing the direct calculation of HTER was shown by
\citet{kozlova-shmatova-frolov:2016:WMT}, who had positive results by predicting
BLEU instead of HTER.
Predicting the HTER score with regards to post-editing operations, such as
re-ordering and lexical choices, has been done by adding the relevant features
in the input \cite{sagemo-stymne:2016:WMT}, whereas \citet{tezcan-EtAl:2015:WMT}
use the word-level quality estimation labels as a feature for predicting the
sentence-level score. To the best of our knowledge, all previous work used a
model to directly predict a single HTER score, in contrast to
\citet{avramidis:2014:W14-33}, which trained one separate model for every HTER
component and used the 4 individual predictions to calculate the final score,
albeit with no positive results. In our work we extend that, by employing a more
elegant machine learning approach that predicts four separate labels for the
HTER components but through a single model.

\section{Methods}
\label{sec:methods}

\subsection{Machine Learning}
\label{sec:methods:ml}

The calculation of HTER is based on the count of 4 components, namely the number
of insertions, deletions, substitutions and shifts (e.g. reordering) that are
required for minimally post-editing a MT output towards the correct translation.
The final HTER score is the total number of editing operations divided by the
number of reference words.

\begin{equation}
	\label{eq:hter}
	\textsc{HTER}=\frac{\textrm{\#insertions} + \textrm{\#dels} + \textrm{\#subs} + \textrm{\#shifts}}{\textrm{\#reference words}}
\end{equation}

\begin{figure}[ht]
	\centering
	\begin{tikzpicture}[shorten >=1pt,thick,scale=0.8, every node/.style={scale=0.8}]
		\tikzstyle{unit}=[draw,shape=circle,minimum size=1.15cm]
		\tikzstyle{hidden}=[draw,shape=circle,fill=black!25,minimum size=1.15cm]
 
		\node[unit](x0) at (0,3.5){$x_1$};
		\node[unit](x1) at (0,2){$x_2$};
		\node at (0,1){\vdots};
		\node[unit](xd) at (0,0){$x_D$};
 
		\node[hidden](h10) at (3,4){$h_1$};
		\node[hidden](h11) at (3,2.5){$h_2$};
		\node at (3,1.5){\vdots};
		\node[hidden](h1m) at (3,-0.5){$h_m$};
		\node[unit](y1) at (6,4){$y_1$};
		\node[unit](y2) at (6,2.5){$y_2$};
		\node[unit](y3) at (6,1.0){$y_3$};
		\node[unit](y4) at (6,-0.5){$y_4$};
 
 		\draw[->] (x0) -- (h10);
		\draw[->] (x0) -- (h11);
		\draw[->] (x0) -- (h1m);
 	
 		\draw[->] (x1) -- (h10);
		\draw[->] (x1) -- (h11);
		\draw[->] (x1) -- (h1m);
 
 		\draw[->] (xd) -- (h10);
		\draw[->] (xd) -- (h11);
		\draw[->] (xd) -- (h1m);
 
		\draw[->] (h10) -- (y1);
		\draw[->] (h10) -- (y3);
		\draw[->] (h10) -- (y2);
		\draw[->] (h10) -- (y4);
 
		\draw[->] (h11) -- (y1);
		\draw[->] (h11) -- (y3);
		\draw[->] (h11) -- (y2);
		\draw[->] (h11) -- (y4);
 
		\draw[->] (h1m) -- (y1);
		\draw[->] (h1m) -- (y2);
		\draw[->] (h1m) -- (y3);
		\draw[->] (h1m) -- (y4);
 
		\draw [decorate,decoration={brace,amplitude=8pt},xshift=-4pt,yshift=0pt] (-0.5,4) -- (0.75,4) node [black,midway,yshift=+0.6cm]{features};
		\draw [decorate,decoration={brace,amplitude=8pt},xshift=-4pt,yshift=0pt] (2.5,4.5) -- (3.75,4.5) node [black,midway,yshift=+0.6cm]{hidden layer};
		\draw [decorate,decoration={brace,amplitude=8pt},xshift=-4pt,yshift=0pt] (5.5,4.5) -- (6.75,4.5) node [black,midway,yshift=+0.6cm]{labels};
	\end{tikzpicture}
	\caption{Network graph for the multi-layer perceptron which given the features
	$x_{1\dots D}$ can jointly predict the amount of the post-editing operations
	$y_{1\dots4}$}
	\label{fig:perceptron}
\end{figure}
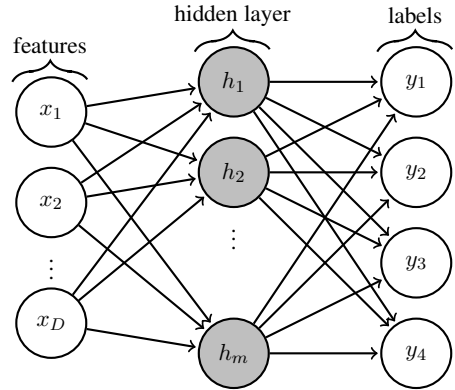

We are here testing 4 different approaches to the prediction of HTER:
\begin{enumerate}
  \item {\bf Baseline with single score}: the baseline system of the
  WMT17 shared task using SVM regression
  \cite{basak2007support} to directly predict the HTER score.
  \item {\bf Combination of 4 SVM models} (4$\times$SVM): this is following
  \citet{avramidis:2014:W14-33} so that it produces 4 separate SVM regression
  models that predict the amount of post-editing operations
  (insertions, deletions, substitutions and shifts respectively). Then,
  HTER is computed based on the 4 separate outputs (Equation~\ref{eq:hter}).
  \item {\bf Single-output perceptron} (MLP): a multi-layer perceptron
  is trained to predict the HTER score
  \item {\bf Multi-output perceptron} (MLP4): a multi-layer perceptron is
  trained given the feature set in the input and the counts of the 4
  post-editing operations as the output labels. Similar to 4$\times$SVM, the separate predictions are used to compute the HTER score
  (Equation~\ref{eq:hter}). The perceptron is depicted in
  Figure~\ref{fig:perceptron}.
\end{enumerate}

%TODO: fix neural network figure 

In the fist two models, SVM follows the baseline set-up of the WMT17 shared
task, using SVM regression with an RBF kernel. The hyperparameters of all three
models, including the number of the hidden units of the perceptron, are tuned
via grid search on cross-validation with 5 folds over the training data.

\subsection{Normalization of predictions}
\label{sec:methods:normalization}

Additionally to the separate models, we are testing here some additional
normalization on the predicted number of post-editing operations, before it is
used to calculate HTER:

\begin{enumerate}[label=\roman*.]
  \item {\bf Integer rounding}: although the model was trained using only
  integers as labels, the regression model resulted into predictions including
  decimals. By assuming that only an integer amount of post-editing operations
  should be valid, we round up the post-editing operations to the closest
  integer.
  \item {\bf Trim negative and big numbers}: MLP4 may also predict numbers
  outside the valid integer range, e.g. providing negative numbers or numbers
  higher than the amount of actual words in the sentence, particularly when
  features have not been normalized.
  Here, we trim the invalid values to the nearest integer within the valid
  range.
\end{enumerate}

\subsection{Optimization measure}
\label{sec:methods:optimization}

Preliminary experiments indicated that the performance of the MLP4
may vary depending on the optimization metric used for tuning the
hyperparameters in a grid search with cross-validation. We tested the
optimization scoring the folds with R$^2$ and Pearson's rho (which is the
official metric of the shared task) in three variations:
\begin{enumerate}[label=\alph*)]
  \item the R$^2$ of the predicted amount of post-editing operations against
  the golden amount of the post-editing operations 
  \item the product of 4 rhos (rho edits); each rho evaluating the predictions
  for one type of post-editing operation (no normalization of predictions) against the golden
  amount of edits for the same post-editing operation 
  \item the rho over the final computed HTER (rho~HTER) against the golden HTER
  without any prior normalization of predictions 
\end{enumerate}

\section{Experiments}
\label{sec:experiment}

Here we present the experiments of testing the various machine learning
approaches on the development set. After the decisions were taken based on the
development set, the models were also applied on the test-sets and the
respective results are also given.
The performance of the models is measured with Pearson's rho, as this is the
basic metric of the WMT17 Shared Task.
A test on statistical significance for comparisons is performed with bootstrap
re-sampling over 1000 samples.
The 4 types of post-editing operations were re-computed with TERCOM on the exact
way that the workshop organizers computed the HTER scores.\footnote{TERCOM ver.
0.7.25 was downloaded from \url{http://www.cs.umd.edu/~snover/tercom}. The
scripts used for running the experiments can be found at
\url{https://github.com/lefterav/MLP4}.}

Similar to the baseline, features values are standardized by removing the mean
and scaled to unit variance. Since the experiment is focusing on machine
learning, for German-English only the baseline features are used.
For English-German, we additionally performed preliminary experiments with the
feature-set from \citet{pub9044} including 94 features that improved QE
performance for translating into German, generated with the software Qualitative
\cite{pub8768}.
The addition of these features did not result into any improvements, so we are
not reporting their results during the development phase (see
Section~\ref{sec:submission} for more details).
The code for training quality estimation models was based on the software
Quest++
\cite{specia-paetzold-scarton:2015:ACL-IJCNLP-2015-System-Demonstrations} and
Scikit-learn \cite{scikit-learn} ver. 1.18.

The development results concerning the presented methods are given below in
this section. The model approaches are tested for both language directions,
whereas the experiments on the normalization of the predictions and ML
optimization are run only for German-English and these observations are applied 
to the models for English-German.

\subsection{Best ML method}

The results concerning the choice of the ML method applied on German-English are
shown in Table~\ref{tab:de-en}.
\begin{table}[ht] \center
\begin{tabular}{lll}
\toprule
method 	& \multicolumn{1}{c}{dev}	& \multicolumn{1}{c}{test} \\
\midrule
SVM		& 0.400 	& 0.441 	\\
4$\times$SVM	& 0.392 	& 0.409 	\\
MLP     & 0.447* 	& 0.447 	\\
MLP4    & 0.476* 	& 0.475** 	\\
\bottomrule
\end{tabular}
\caption{Pearson rho correlation against golden labels concerning the 4
different approaches for predicting HTER for German-English. (*)
indicates significant improvement ($\alpha=0.05$) over the SVM baseline (**)
significant improvement over all models } 
\label{tab:de-en}
\end{table}

The approach of MLP4 achieves a small but significant improvement over the
baseline and the 4$\times$SVM on the development set.
On the development set both MLP and MLP4 beat significantly the baseline, but
MLP4 is not significantly better than MLP.
Nevertheless, when applied to the test-set, the improvement achieved with MLP4
is significant as compared to all other ML methods.

\begin{table}[ht]
\center
\begin{tabular}{llrc}
\toprule
%feature-set & \multicolumn{2}{c}{baseline} && \multicolumn{2}{c}{augmented} \\
method 	& \multicolumn{1}{c}{dev}	& \multicolumn{1}{c}{test2017}	& test2016	\\
\midrule
SVM				& ~0.414 & 0.402~~	  & 0.407	\\
4$\times$SVM	& ~0.049	& -0.071~~  &	0.044 \\
MLP     		& ~0.343  & 0.335~~  & 0.327 \\
MLP4   			& ~0.429*  & 0.412~~  & 0.412 \\
\bottomrule
\end{tabular}
\caption{Performance of the 4 different approaches for predicting HTER for
English-German
(*) indicates a significant improvement ($\alpha=0.1$)
over the baseline 
}
\label{tab:en-de}
\end{table}

The same approaches show moderate improvements when applied to English-German
with the baseline feature set (Table~\ref{tab:en-de}).
MLP4 achieves higher correlation score than the baseline, but the difference is
small and it is significant only for the development set. When compared to the
other two methods, though, MLP4 achieves a significant improvement.
In contrast to the direction German-English, in English-German the MLP with one
output performs worse than the baseline.
4$\times$SVM fails to predict HTER as its predictions achieve zero correlation.
Since the individual models failed to predict the post-editing operations
separately, this may be an indication that among the 4 post-editing operations
in English-German there are dependencies which are stronger than the ones in
German-English.

\subsection{Normalization of predicted post-editing operations}

The effect of the normalization of the predicted post-editing operations of
MLP4, prior to the calculation of the final HTER score, is shown in
Table~\ref{tab:normalization}.

\begin{table}[ht]
\center
\begin{tabular}{lcc}
\toprule		
				& dev	& test   \\
\midrule
original labels	& 0.473	& 0.471 \\
trim 			& 0.476 & 0.475 \\
round			& 0.456	& 0.469 \\
round \& trim & 0.456 & 0.467 \\
\bottomrule
\end{tabular}
\caption{Performance improvements by introducing rounding and cut-off for the
predicted post-editing operations (German-English)}
\label{tab:normalization}
\end{table}

The experiment indicates some small improvement when we trim the invalid
predicted values, so we use this for all other calculations. Preliminary
experiments indicated more significant improvements when the feature values have
not been standardized and re-scaled prior to the training.

\subsection{ML optimization}

The effect of using different methods for hyperparameter optimization is show in
Table~\ref{tab:optimization}.

\begin{table}[ht]
\center
\begin{tabular}{lcc}
\toprule
			& \multicolumn{1}{c}{dev}	& \multicolumn{1}{c}{test} \\
\midrule
R$^2$ 		& 0.440 & 0.454 \\
rho HTER 	& 0.431 & 0.457  \\
rho edits	& 0.476 & 0.475 \\
\bottomrule
\end{tabular}
\caption{Experimentation with different optimization measures for defining the
perceptron hyperparameters (German-English model) }
\label{tab:optimization}
\end{table}

The product of the 4 rhos, calculated over the 4 types of post-editing
operations (rho edits) has slightly better performance than the other scoring
methods, nevertheless the difference is not statistically significant.
Using these findings just as an indication, we perform all experiments by
optimizing the hyperparameters with \emph{rho edits}.

The optimized hyperparameters for the SVM models are shown in
Table~\ref{tab:hyperparameters}, whereas the ones for the MLP models are shown
in Table~\ref{tab:hyperparameters_mlp}. All SVMs have an RBF kernel and all MLPs
are optimized with \emph{adam} as a solver. It is noteworthy that for German-English a network topology with
multiple hidden layers performed better, which is an indication that the mapping
between features and labels in this language pair is much more complex than the
one for German-English.

\begin{table}[ht]
\center
\begin{tabular}{llcrl}
\toprule
langpair	& model					& $\epsilon$	& C		&	\multicolumn{1}{c}{$\gamma$}	\\
\midrule 
de-en		& SVM					& 0.1			& 10	&	0.001		\\
			& 4$\times$SVM (ins)   & 0.2			& 10	&   0.01 		\\
			& 4$\times$SVM (del)   & 0.2			& 10	&   0.01 		\\
			& 4$\times$SVM (shifts)& 0.2			& 10		&   0.01 		\\
			& 4$\times$SVM (subst) & 0.1			& 10 	&   0.01 		\\
\midrule
en-de		& SVM			& 0.1			& 1		&	0.01		\\
			& 4$\times$SVM (ins)   & 0.2			& 1 	&   0.001 		\\
			& 4$\times$SVM (del)   & 0.1			& 1	    &   0.001 		\\
			& 4$\times$SVM (shifts)& 0.1			& 1		&   0.001 		\\
			& 4$\times$SVM (subst) & 0.2			& 1 	&   0.001 		\\
\bottomrule	
\end{tabular}
\caption{Hyperparameters used after the optimization of the SVM models}
\label{tab:hyperparameters}
\end{table}

\begin{table}[ht]
\center
\setlength\tabcolsep{4pt}
\begin{tabular}{llcccl}
\toprule
langp. & model & act. & $\alpha$ & tol. & hidden units \\
%\rotlangpair} & \rot{model} 	& \rot{activ.} 	& \rot{$\alpha$} &
%\rot{tolerance} & \rot{hidden units} \\ 
\midrule
de-en 	& MLP 	& relu	& 0.10	& $10^{-9}$ & 1: 100		\\	 
		& MLP4  & relu	& 0.10	& $10^{-3}$ & 1: 300 		\\				 
en-de 	& MLP   & tanh 	& 0.01	& $10^{-3}$ & 3: 150, 75, 6 \\
   		& MLP4	& tanh  & 0.10  & $10^{-3}$ & 2: 300, 150	\\
\bottomrule
\end{tabular}
\caption{Hyperparameters and network topology after the optimization of the MLP
models}
\label{tab:hyperparameters_mlp}
\end{table}

\section{Submission and post-mortem analysis}
\label{sec:submission}

Whereas previous sections described a full development phase in order to support
the idea of the multi-output MLP, this section is focusing on our exact
submission for the Quality Estimation Task of WMT17.
Unfortunately, a development issue prior to the submission prevented our
experiments from standardizing the feature values and scaling them to unit
variance. Since the performance of SVM suffers from non-scaled feature values,
this led our development phase to proceed by contrasting with a much lower
baseline than the one finally provided by the workshop organizers. Non-scaled
features and other settings affected also the performance of MLP models, and
therefore the scores on our final submissions are significantly lower than the official baseline. The
issue became apparent only after the submission, so we then re-computed the all
models with standardized and scaled feature values. The results presented in
Section~\ref{sec:experiment} are based on these corrected models.

The submitted models used both rounding and trimming of predicted integers
(Section~\ref{sec:methods:normalization}).
The MLPs were optimized with an $\alpha=0.01$, \emph{tanh} as an activation
function, and \emph{adam} as a solver. The German-English model got optimal with
300 hidden units.
The English-German was trained using the additional 52 features from
\citet{pub9044} which gave good development results only with 3,000 hidden
units, which is an indication of overfitting.

\begin{table}[ht]
\center
\begin{tabular}{llc}
\toprule
method 	& \multicolumn{1}{c}{dev}	& test \\
\midrule
baseline (ours)	    & 0.32	& 0.34	\\
MLP4 (submitted)    & 0.40 	& 0.40  \\
\midrule
baseline (official)	& 0.40	& 0.44	\\
MLP4 (corrected)	& 0.48  & 0.48  \\
\bottomrule
\end{tabular}
\caption{Scores for the submitted models and for their corrected
versions after the submission (German-English)}
\label{tab:submission_de-en}
\end{table}

\begin{table}[ht]
\center
\begin{tabular}{lccc}
\toprule
method 	& \multicolumn{1}{c}{dev}	& test2017 & test2016 \\
\midrule
baseline (ours)			& 0.19	& 0.20 & 0.12  \\
MLP4 (submitted)		& 0.40 	& 0.11 & 0.13 \\
\midrule
baseline (official)		& 0.41	& 0.40 & 0.40 \\
MLP4 (corrected)		& 0.43  & 0.41 & 0.41 \\
\bottomrule
\end{tabular}
\caption{Scores for the submitted models and for their corrected versions after
the submission English-German}
\label{tab:submission_en-de}
\end{table}

A comparison of the models developed before the submission and the corrected
ones are shown in Tables \ref{tab:submission_de-en} and
\ref{tab:submission_en-de}. The submitted model for German-English was expected
to be significantly better than the baseline, whereas the one for English-German
with the additional features had strong indications of overfitting and performed
indeed poorly at the final test-sets. 

The corrected models perform better after scaling is added and the rounding of
integers is disabled. The corrected model for English-German converges without
overfitting after removing the additional features and adding one more hidden
layer. These models, if submitted to the shared task, despite comparing with the
baseline, they would still score lower than almost all the others submitted
methods. Though, we need to note that this should still be satisfactory, as we
did not perform any feature engineering, aiming at confirming our hypothesis for using
multi-output models.

\section{Conclusion and further work}

In this submission we investigated the idea of using a multi-layer perceptron
in order to jointly predict the 4 distinct post-editing operations, which are
then used for calculating the HTER score. The experiments show some small but
significant improvements on both the development set and the test-set for
German-English, but the same approach showed improvement only on the development
set when applied English-German.

Despite not having conclusive results yet, we think that the idea is promising
and that further experiments could have positive impact. Concerning the current
development, several issues need to be further investigated, such as possible
ways to avoid the lack of robustness of the perceptron.
Since this work did not focus at feature engineering, further work could profit
from introducing features highly relevant to the specific types of post-editing
operations, or even upscaling observations from word-level and phrase-level QE.
On the machine-learning level, additional hidden layers and more work on the
number of hidden units might be of benefit. Finally, evaluation specific to the
types of the predicted post-editing operations could provide hints for further
improvement.

\label{sec:conclusions}

\section*{Acknowledgments}
This work has received funding from the European Union's Horizon 2020
research and innovation program under grant agreement N$^{o}$ 645452 (QT21).

% \bibliography{emnlp2017}
\bibliographystyle{emnlp_natbib}
\bibliography{library}

\end{document}